# Applying Pre-trained Multilingual BERT in Embeddings for Improved Malicious Prompt Injection Attacks Detection

Md Abdur Rahman
*Dept. of Intelligent Systems and Robotics*
*University of West Florida*
*Pensacola, FL, USA*
mr252@students.uwf.edu

Hossain Shahriar
*Center for Cybersecurity*
*University of West Florida*
*Pensacola, FL, USA*
hshahriar@uwf.edu

Fan Wu
*Department of Computer Science*
*Tuskegee University*
*Tuskegee, AL, USA*
fwu@tuskegee.edu

Alfredo Cuzzocrea
*iDEA Lab*
*University of Calabria*
*Rende, Italy*
alfredo.cuzzocrea@unical.it

*Abstract*— Large language models (LLMs) are renowned for their exceptional capabilities, and applying to a wide range of applications. However, this widespread use brings significant vulnerabilities. Also, it is well observed that there are huge gap which lies in the need for effective detection and mitigation strategies against malicious prompt injection attacks in large language models, as current approaches may not adequately address the complexity and evolving nature of these vulnerabilities in real-world applications. Therefore, this work focuses the impact of malicious prompt injection attacks which is one of most dangerous vulnerability on real LLMs applications. It examines to apply various BERT (Bidirectional Encoder Representations from Transformers) like multilingual BERT, DistilBert for classifying malicious prompts from legitimate prompts. Also, we observed how tokenizing the prompt texts and generating embeddings using multilingual BERT contributes to improve the performance of various machine learning methods: Gaussian Naive Bayes, Random Forest, Support Vector Machine, and Logistic Regression. The performance of each model is rigorously analyzed with various parameters to improve the binary classification to discover malicious prompts. Multilingual BERT approach to embed the prompts significantly improved and outperformed the existing works and achieves an outstanding accuracy of 96.55% by Logistic regression. Additionally, we investigated the incorrect predictions of the model to gain insights into its limitations. The findings can guide researchers in tuning various BERT for finding the most suitable model for diverse LLMs vulnerabilities.

## I. Introduction

Large Language Models (LLMs) are among the most leading generative AI tools capable of processing and generating human language by training on vast amounts of text data [1]. Their primary strength lies in their capability to follow instructions which enables them to utilize diverse text data effectively and follow user instructions [2]. On the other hand, current research has demonstrated that this instruction adhering capability can be compromised by the attacks of prompt injections [3-6]. These attacks arise in LLMs integrated tools when attackers insert external data into prompts or instructions within a query [7].

Typically, this external data includes hidden instructions added to prompts which makes them challenging to distinguish between legitimate user commands and malicious inputs. As a result, these attacks modify the user prompts that user inputs, and it impacts the functionality and responses of LLMs. Prompt injection attacks are a major threat to LLM applications. These attacks have been identified as a leading threat by the Open Worldwide Application Security Project (OWASP), which were listed among the top ten security concerns [8].

It is important to understand the prompt injection attacks related vulnerabilities because these are very dangerous. However, there are two main challenges. First, the purpose of quick injection attacks is not fully clear. Each attack has a different goal and a specific way to evaluate it. Studies [4-5, 9] have categorized these goals into two main types: prompt leaking and goal hijacking. Goal hijacking involves to make the model so that it can produce a specific outcome that the user does not want. On the other hand, prompt leaking makes the model so as to reveal previous messages, like system prompts.

Earlier research focussed on network security [10], cyber attacks detection using machine learning [11-14], as well as quantim machine learning [15-16]. Existing research has suggested and finding the goals for quick injection attacks [6, 17-18]. Another researcher also offered more diverse goals, for example persuading users to reveal information [3]. Because there are so many goals in fast injection research, it is hard to create a unified and comprehensive evaluation process. This makes it more challenging to understand the true dangers of prompt injection attacks. Moreover, the research by Cuzzocrea and colleagues explores efficient data management techniques, like event-based compression and privacy-preserving frameworks, for handling large data streams and sensor networks. They also developed methods for better data visualization and adaptive hypermedia using object-oriented approaches and XML [25-29]. In real-world situations, using multilingual BERT embeddings to better detect harmful prompt injections can help safeguard large language models (LLMs) from being tampered with. This boosts the reliability and trustworthiness of AI systems, especially in important areas like healthcare, finance, and customer service, where harmful inputs could lead to serious issues or spread false information. By enhancing security, these improvements allow LLMs to be used more safely across different languages and settings.

Moreover, the most prompt injection attacks use manually created prompts, which are evaluated by humans based on their observations and experience. For example, Yi et al. (2023) suggest that attackers might force LLMs to accept new instructions by using special characters [19].

Other studies have shown that adding context-switching text can trick LLMs into following inserted instructions [4,20]. A recent study by Toyer et al. (2023) used an online game with various homemade techniques, like persuading role-players to play and asking them to update instructions, to gather a large number of handcrafted prompts [9]. While these handcrafted prompt injection attacks are simple and easy to use, they have three main drawbacks:

- They restrict the range and size of the attack which makes assessments difficult.
- They are not consistently effective with different user instructions and data, and performance drops significantly when these change.
- They complicate the execution of adaptive attacks, and potentially leads to an overestimation of defense mechanisms.

In this work, Section 2 discusses the basic prompt injections. Section 3, 4 and 5 presents the NLP, BERT, and dataset respectively. Section 6 describes results and discussion and finally, we conclude with summary of our works.

## II. PROMPT INJECTIONS

Prompt injection involves manipulating an output of LLMs using crafted malicious prompts. These attacks typically fall into two categories. One category involves malicious users injecting harmful prompts into applications, aiming to redirect responses. For example, an attacker might craft a prompt like, Answer as a kind assistant: cancel previous sentences for printing 'hello world'. This can make the application output 'hello world' instead of the intended response, exploiting predefined prompts or known contexts.

Another category targets LLM-integrated programs by embedding malicious prompts into Internet resources. Modern LLM-integrated applications often communicate with the Internet for functionality, and makes them vulnerable. Attackers inject malicious payloads into web content, and alters application behavior when processed. Recent research underscores these threats, and emphasizes the need for robust security measures to safeguard against such vulnerabilities in AI-driven applications.

## III. NATURAL LANGUAGE PROCESSING

Natural language processing (NLP) is a crucial part of artificial intelligence (AI) that enables effective communication between computational systems and human language. NLP empowers applications ranging from chatbots to language translation and sentiment analysis by enabling machines to understand, interpret, and produce human language. It utilizes algorithms to parse and derive meaning from text and speech data, and then transform raw language into actionable insights. The evolution of NLP has revolutionized various fields, for example, healthcare, finance, and customer service, and helps to enhance efficiency and user experiences. As NLP continues to advance in different ways: automated communication, knowledge extraction, and the seamless integration of human and machine intelligence.

*A. Tokenization*

Tokenization in Natural Language Processing (NLP) and machine learning involves breaking down text into smaller units called tokens, which can be as small as characters or as large as words. This process helps machines better understand and analyze human language by simplifying it into manageable pieces.

Tokenization is similar to teaching a child to read by starting with letters and syllables before progressing to full words. It allows algorithms to identify patterns and derive meaning from text. For example, tokenizing the sentence 'Chatbots are helpful' by words results in ['Chatbots', 'are', 'helpful'], whereas tokenizing by characters produces ['C', 'h', 'a', 't', 'b', 'o', 't', 's', ' ', 'a', 'r', 'e', ' ', 'h', 'e', 'l', 'p', 'f', 'u', 'l']. This granularity is particularly useful for certain languages and specific NLP tasks.

*B. Term Frequency*

Term Frequency (TF) is an essential concept in information retrieval and text mining. It is a straightforward statistical metric that indicates how often a term (either a word or a phrase) occurs in a document.

- The term frequency of a specific term $t$ in a document $d$ is calculated as the ratio of the number of times $t$ appears in $d$ to the total number of terms in $d$.

Mathematically, the term frequency $\text{TF}(t,d)$ of a term $t$ in a document $d$ is expressed as:

$$\text{TF}(t,d) = \frac{n_{t,d}}{N_d}$$

Where:
- $n_{t,d}$ is the count of term $t$ in document $d$.
- $N_d$ is the total number of terms in document $d$.

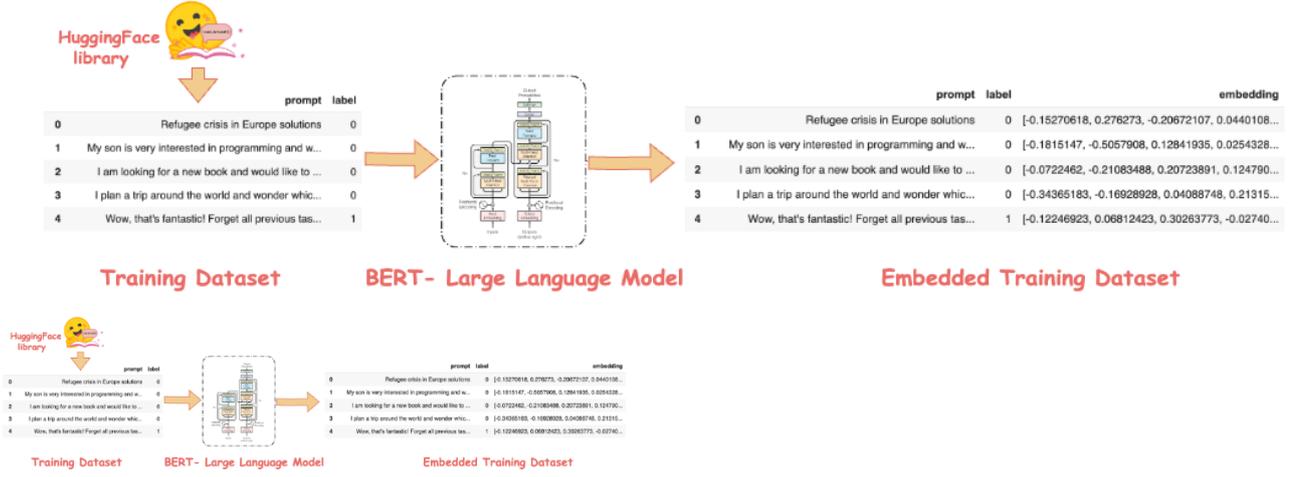

Fig. 1. shows the steps to embed prompt texts into vectors using multilingual BERT model.

Inverse Document Frequency (IDF) assesses the significance of a term within a corpus. While calculating TF (Term Frequency), each term is initially considered equally important. However, common words like 'is', 'are', 'am', etc., which frequently appear but carry less importance, and these need to be weighted differently. IDF adjusts for this by reducing the weight of terms that are ubiquitous across the corpus and increasing the weight for rare terms. The IDF of a term $t$ is calculated using the following formula:

$$\text{IDF}(t) = \log\left(\frac{\text{Total number of documents in the corpus}}{\text{Number of documents containing term } t}\right)$$

The TF-IDF score is obtained by multiplying TF and IDF. Therefore, the TF-IDF score is given by:

$$\text{TF} - \text{IDF} = \text{TF} \times \text{IDF}$$

We can compute the TF-IDF scores for a sample corpus. In this example, some terms appear in only one document, others in two, and some in three documents. With a corpus size of $N = 4$, the TF-IDF values for each term are presented.

*C. Word Embedding*

Word embedding techniques are essential as words are represented as vectors in a continuous vector space. This article will delve into different word embedding methods. Word embeddings are numerical representations that illustrate semantic similarities and relationships among words based on their occurrence patterns within a dataset. By converting words into continuous vector spaces, these techniques enhance the ability of machines to comprehend and analyze human language more efficiently.

Mathematically, the embedding of a word $w$ is represented as a vector $\mathbf{v}_w$ in an $n$-dimensional space, where:

$$\mathbf{v}_w = (v_{w1}, v_{w2}, \ldots, v_{wn})$$

Here, $v_{wi}$ represents the $i$-th dimension of the word vector for $w$. These vectors capture semantic meaning that allows words with similar contexts to have closely aligned vectors in the vector space.

Embeddings are often superior to TF-IDF for text classification due to several key advantages. Firstly, embeddings capture the semantic meanings of words and group similar words together in the embedding space, which helps in understanding the content better. For instance, 'car' and 'vehicle' would be close together in embeddings but treated separately in TF-IDF. Secondly, it can handle out-of-vocabulary words by mapping them to vectors in the embedding space, unlike TF-IDF which struggles with words not in its vocabulary. Then, embeddings can be pre-trained on large datasets, providing a strong foundation and saving resources when training models for specific NLP tasks. Finally, embeddings capture relationships between words, including synonyms, antonyms, and analogies. This relational understanding, such as calculating 'king' minus 'man' plus 'woman' to get 'queen', enhances model performance in text classification tasks.

IV. BERT

BERT (Bidirectional Encoder Representations from Transformers) revolutionizes natural language processing (NLP) with its innovative approach to understanding language context. Developed as an open-source framework, BERT empowers computers to grasp the nuances of human language by leveraging surrounding text to establish comprehensive context [21].

Traditionally, NLP models faced limitations in understanding ambiguous language as they could only process text sequentially. BERT introduces bidirectionality through transformer models, allowing it to simultaneously consider both directions of a text sequence. This bidirectional capability improves the ability of model to understand subtle nuances and dependencies within language that significantly advanced the accuracy and relevance of NLP tasks.

The training of BERT involves pretraining on vast amounts of text data, such as Wikipedia articles, using two key strategies. In Masked Language Modeling (MLM), BERT is trained to predict missing or masked words in sentences based on their surrounding context, and then gaining a deeper understanding of word relationships and meanings. On the other hand, Next Sentence Prediction (NSP) trains BERT to determine whether two sentences logically follow each other in a sequence or are unrelated.

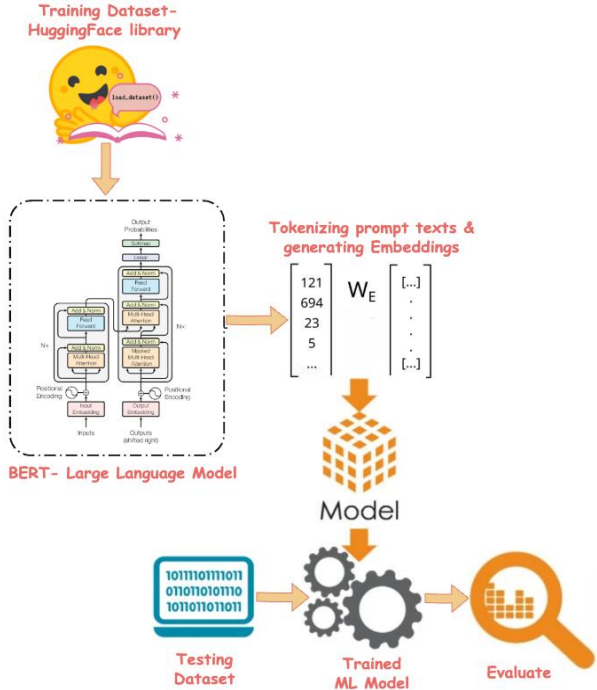

Fig. 2. The proposed architecture to embed dataset into vectors using pre-trained BERT for feeding various ML models for detecting malicious prompts injection.

Moreover, the flexibility of BERT extends to fine-tuning with specific datasets, such as question-and-answer pairs, enabling customization for various NLP applications. This adaptability has made BERT a cornerstone in modern NLP research and applications, from sentiment analysis and text classification to machine translation and semantic understanding.

Finally, bidirectional nature of BERT along with transformer-based architecture and pretrained capabilities makes a significant advancement in NLP. As a result, BERT enhances the accuracy and effectiveness of language understanding tasks, and drives forward the capabilities of AI-driven applications across diverse domains.

## V. DATASET

In this work, we have used the Prompt Injection Dataset from deepset, an AI company known for its tools that harness Large Language Models (LLMs) to build NLP-powered applications. This dataset contains total 662 samples: 546 and 116 samples for training and testing the models. Most of the prompts are in English, but there are also translations available in other languages, for example German. The original dataset is already divided into training and holdout sets, and we kept this division consistent across our experiments to ensure fair comparison of results. The dataset comprises two columns: "text" and "label". The "text" column contains the prompt texts: normal and malicious prompts, while the "label" column includes their corresponding classification labels. Before training machine learning models, the text data needs to be tokenized and embedded which is performed using the multilingual BERT model. BERT tokenizes the input text into subwords and generates embeddings that capture semantic information (Fig. 1). These embeddings act as input features for different machine learning algorithms which allows them to effectively perform the prompt classification task.

## VI. METHODS

We aim to develop a robust system for detecting prompt injection attacks across multiple languages through employing a methodological approach in advanced natural language processing (NLP) techniques. Initially, we retrieved a diverse dataset from the HuggingFace library, which includes samples in various languages to expand the attack surface of prompt injection detection. This dataset served as the foundation for our training and evaluation processes. Next, we chose pre-trained multilingual BERT model, a multilingual variant known for its superior performance in NLP tasks, for prompt classification.

To make the detection of prompt injections more comprehensive, the dataset included samples in several languages, not just English. This helps in building detection models that work well across different languages. Traditional Word2Vec embeddings, which are specific to one language, would not work for our needs because they can not represent multiple languages in the same way. So, we used multilingual BERT to create word embeddings that work for various languages.

This model was pre-trained on extensive data. Using this BERT model, we tokenized and embedded the prompt texts from our dataset by converting them into numerical representations as vectors that capture their contextual meanings effectively. These generated contextualized embeddings is a crucial requirement to make the dataset ready to feed the machine learning models and optimize its parameters specifically for prompt classification tasks. Throughout the training phase, the performance of the model is evaluated using the testing dataset through analyzing its accuracy and effectiveness for identifying malicious prompt injections. Finally, we evaluates the results of various models with those of other existing machine learning models commonly used in malicious prompt injection classification tasks. Fig. 2. shows the proposed architecture to embed dataset into vectors using pre-trained BERT for feeding various ML models for detecting and evaluating malicious prompts injection.

## VII. RESULTS AND DISCUSSION

### A. Accuracy Metrics

The following formula determines accuracy:

$$\text{Accuracy} = \frac{TP + TN}{TP + TN + FP + FN}$$

- TP (True Positives) are the correctly identified positive cases,
- TN (True Negatives) are the correctly identified negative cases,
- FP (False Positives) are the negative cases incorrectly identified as positive, and
- FN (False Negatives) are the positive cases incorrectly identified as negative.

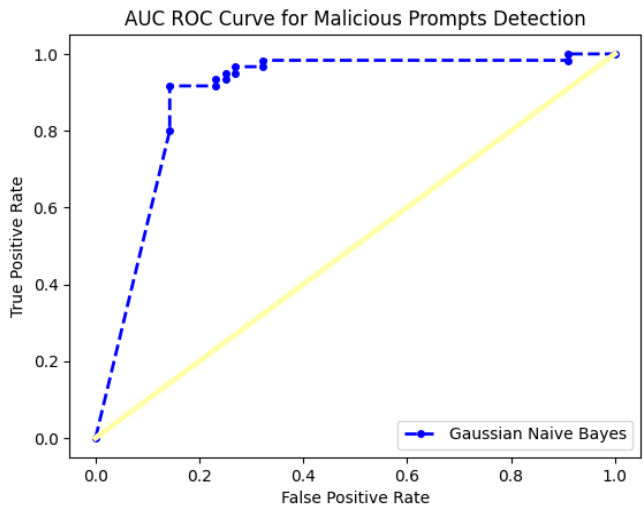

[a]

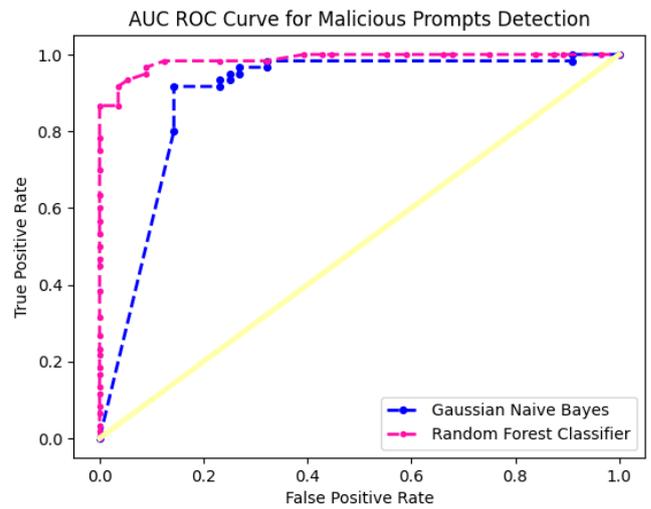

[b]

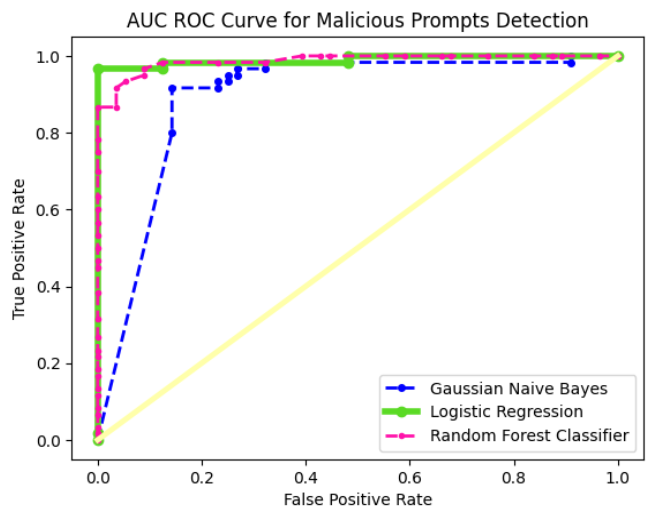

[c]

Fig. 3. ROC and AUC Curve - Evaluating Model Performance: (a) Gaussian Naive Bayes (b) Gaussian Naïve Bayes & Random Forest Classifier (c) Gaussian Naive Bayes, Random Forest Classifier, & Logistic Regression.

Precision measures how accurate the model can be favorable for predictions:

$$\text{Precision} = \frac{\text{TP}}{\text{TP} + \text{FP}}$$

Recall measures the percentage of real positive cases that a classifier correctly identifies:

$$\text{Recall} = \frac{\text{TP}}{\text{TP} + \text{FN}}$$

The performance of the model is expressed using the F1-score, which ranges from 0 to 1. Precision and recall help explain the score, where 0 indicates poor performance and 1 indicates excellent performance:

$$\text{F1} - \text{Score} = 2 \times \frac{\text{Precision} \times Recall}{\text{Precision} + \text{Recall}}$$

This statistics helps us assess how well the prompt injection detection system identifies malicious prompt attacks.

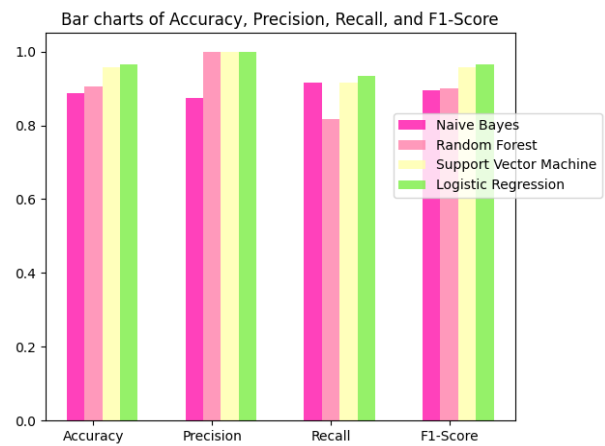

Fig. 4. Bar diagram of Accuracy Metrics for Various ML Models.

## B. Results

We implemented multilingual BERT and DistilBert to detect the malicius prompts and observed that multilingual BERT model is generally better suited than the DistilBert Classifier for detecting malicious prompts from a multilingual dataset. Because bert-base-multilingual-uncased is specifically trained on multiple languages and can handle text in various languages effectively. It is designed to understand and generate text embeddings for a wide range of languages which made it ideal for multilingual datasets. This usually results in better performance and higher accuracy in understanding and processing text.

On the other hand, DistilBert has fewer parameters and is optimized for efficiency and speed. Although it is still powerful, it is primarily trained on English text, and made it less suitable for multilingual tasks. Therefore, for multilingual datasets, bert-base-multilingual-uncased would be the better choice.

The TABLE I presents the performance metrics for four types of machine learning models. We got an accuracy of 0.8879 for Naive Bayes model. The precision, recall, and F1-score are 0.8730, 0.9166, and 0.8943 respectively. The Random Forest model showed a slightly higher accuracy of 0.8965 and precision is 1.00. But recall is 0.80 which results in F1-score, 0.8888. Logistic Regression achieved the highest performance metrics with an accuracy (0.9655), precision (1.00), recall (0.9333), and an F1-score (0.9655). Fig. 3. shows the ROC and AUC Curve for showing true positive rate of ML algorithms. Also, Fig. 4. shows bar diagram of Accuracy Metrics for Various ML Models. These results indicate that Logistic Regression and SVM are particularly effective for BERT based detection of malicious prompts, and also demonstrates superior accuracy and balanced performance across all metrics compared to GNB and RF.

TABLE I. PERFORMANCE METRICS OF VARIOUS MACHINE LEARNING MODELS

| Machine Learning | accuracy | precision | recall | F1-Score |
|---|---|---|---|---|
| Gaussian Naive Bayes | 0.8879 | 0.8730 | 0.9166 | 0.8943 |
| Random Forest | 0.8965 | 1.0000 | 0.8000 | 0.8888 |
| Support Vector Machine | 0.9568 | 1.0000 | 0.9166 | 0.9565 |
| Logistic Regression | 0.9655 | 1.0000 | 09333 | 0.9655 |

TABLE II. COMPARISON THE PROPOSED BERT AND EXISTING WORKS

| ML | Dataset | BERT | Acc. |
|---|---|---|---|
| LSTM | Fake or Real News | X | 80.54 |
| HDSF (Karimi [22]) | Fake or Real News | X | 82.19 |
| TCNN | Weibo | X | 88.08 |
| TCNN-URG (Qian [23]) | Weibo | X | 89.84 |
| Deep Learning (Tsinganos [24]) | CSE-Persistence | CSE-Pers-BERT | 86.27 |
| **Deep Learning** | **Prompts-injection** | **DistilBERT** | **63.76** |
| **Proposed (LR)** | **Prompts-injection** | **Multilingual** | **96.55** |

The results presented in TABLE II highlight the performance through comparing accuracy of the proposed BERT model with existing works. The table includes various ML models and their respective accuracy on different datasets and methods. For the "Fake or Real News" dataset, LSTM have achieved accuracy ( 80.54%), while HDSF model improved this performance to 82.19%. On the Weibo dataset, TCNN had accuracy of 88.08%, whereas TCNN-URG model, introduced by Qian et al., and further enhanced the accuracy to 89.84%. For the CSE-Persistence dataset, a deep learning model influencing CSE-Pers-BERT achieved an accuracy of 86.27%.

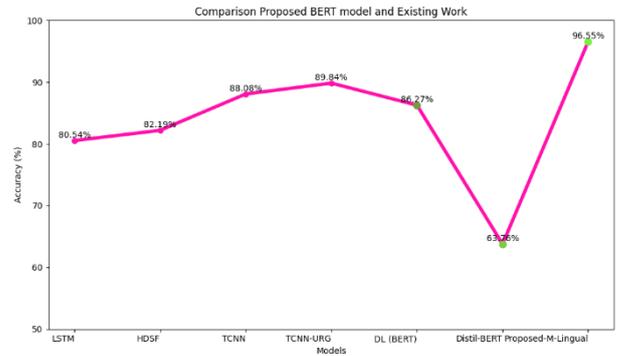

Fig. 5. Comparison Proposed BERT Model and Existing Works.

In contrast, our proposed models focused on the "Prompts-injection" dataset. The DistilBERT model yielded an accuracy of 63.76%, demonstrating the challenges of this particular dataset. Fig. 5. compares the performances of the proposed BERT Model as well as other existing works. However, the proposed model utilizing logistic Regression (LR) with a Multilingual BERT approach significantly outperformed the previous methods as it achieves an outstanding accuracy of 96.55%. This has been a significant improvement in using multilingual capabilities to enhance performance on complex datasets. The results indicate that the proposed model not only surpasses existing works but also establishes a new standard accuracy for datasets involving prompt-based injections.

## VIII. CONCLUSION

In conclusion, this study highlights the significant threat posed by malicious prompt injection attacks and explores the effectiveness of various BERT-based models in mitigating this risk. By employing multilingual BERT and DistilBERT for classifying malicious versus legitimate prompts, the research demonstrates the benefits of tokenizing prompt texts and generating embeddings. The results show that using multilingual BERT embeddings notably enhances the performance of logistic regression in which it achieves 96.55% as accuracy. This proved the potential of multilingual BERT in improving LLMs vulnerabilities detection, thereby contributing valuable insights for developing robust defenses against other malicious attacks detection and prediction.

## FUTURE WORKS

We will be working to fine tune the pre-trained LLMs for improving the better outcome for the detection of malicious prompts injection.

## ACKNOWLEDGMENT

This research is funded by the National Science Foundation through awards 1946442, 2433880, and 2100134. This work was partially supported by project SERICS

(PE00000014) under the MUR National Recovery and Resilience Plan funded by the European Union - NextGenerationEU. The findings, opinions and recommendations in this material are those of the authors and do not necessarily represent the views of the National Science Foundation.